\documentclass[11pt,a4paper]{article}
\usepackage[hyperref]{acl2021}
\usepackage{times}
\usepackage{arydshln}
\usepackage{booktabs}
\usepackage{float}
\usepackage{latexsym}
\usepackage{tabularx}
\usepackage[ruled, vlined, linesnumbered]{algorithm2e}
\usepackage{setspace}
\usepackage{subfigure}   
\usepackage{multirow}
\usepackage{graphicx}
\usepackage{amsfonts}
\usepackage{amsmath}

\usepackage{microtype}

\aclfinalcopy 
\def\aclpaperid{2551} 

\title{Dialogue Graph Modeling for Conversational Machine Reading}

\author{Siru Ouyang\textsuperscript{1,2,3,*}, Zhuosheng Zhang\textsuperscript{1,2,3,\thanks{\ \ Equal contribution. $\dagger$Corresponding author. This paper was partially supported by National Key Research and Development Program of China (No. 2017YFB0304100), Key Projects of National Natural Science Foundation of China (U1836222 and 61733011), Huawei-SJTU long term AI project, Cutting-edge Machine Reading Comprehension and Language Model. This work was supported by Huawei Noah's Ark Lab.} }, Hai Zhao\textsuperscript{1,2,3,$\dagger$} \\
\textsuperscript{1} Department of Computer Science and Engineering, Shanghai Jiao Tong University\\
\textsuperscript{2} Key Laboratory of Shanghai Education Commission for Intelligent Interaction\\
and Cognitive Engineering, Shanghai Jiao Tong University, Shanghai, China\\
\textsuperscript{3}MoE Key Lab of Artificial Intelligence, AI Institute, Shanghai Jiao Tong University, Shanghai, China\\
\texttt{\{oysr0926,zhangzs\}@sjtu.edu.cn,zhaohai@cs.sjtu.edu.cn}\\
}

\date{}

\begin{document}
\maketitle
\begin{abstract}
Conversational Machine Reading (CMR) aims at answering questions in complicated interactive scenarios. Machine needs to answer questions through interactions with users based on given rule document, user scenario and dialogue history, and even initiatively asks questions for clarification if necessary. 
Namely,  the answer to the task needs a machine in the response of either \textsl{Yes, No, Irrelevant} or to raise a follow-up question for further clarification. To effectively capture multiple objects in such a challenging task, graph modeling is supposed to be adopted, though it is surprising that this does not happen until this work proposes a dialogue graph modeling framework by incorporating two complementary graph models, i.e., explicit discourse graph and implicit discourse graph, which respectively capture explicit and implicit interactions hidden in the rule documents.
The proposed model is evaluated on the ShARC benchmark and achieves new state-of-the-art by first exceeding the milestone accuracy score of 80\%. The source code of our paper is available at \url{https://github.com/ozyyshr/DGM}
\end{abstract}

\section{Introduction}

\begin{figure}[ht]
\centering
\includegraphics[width=0.5\textwidth]{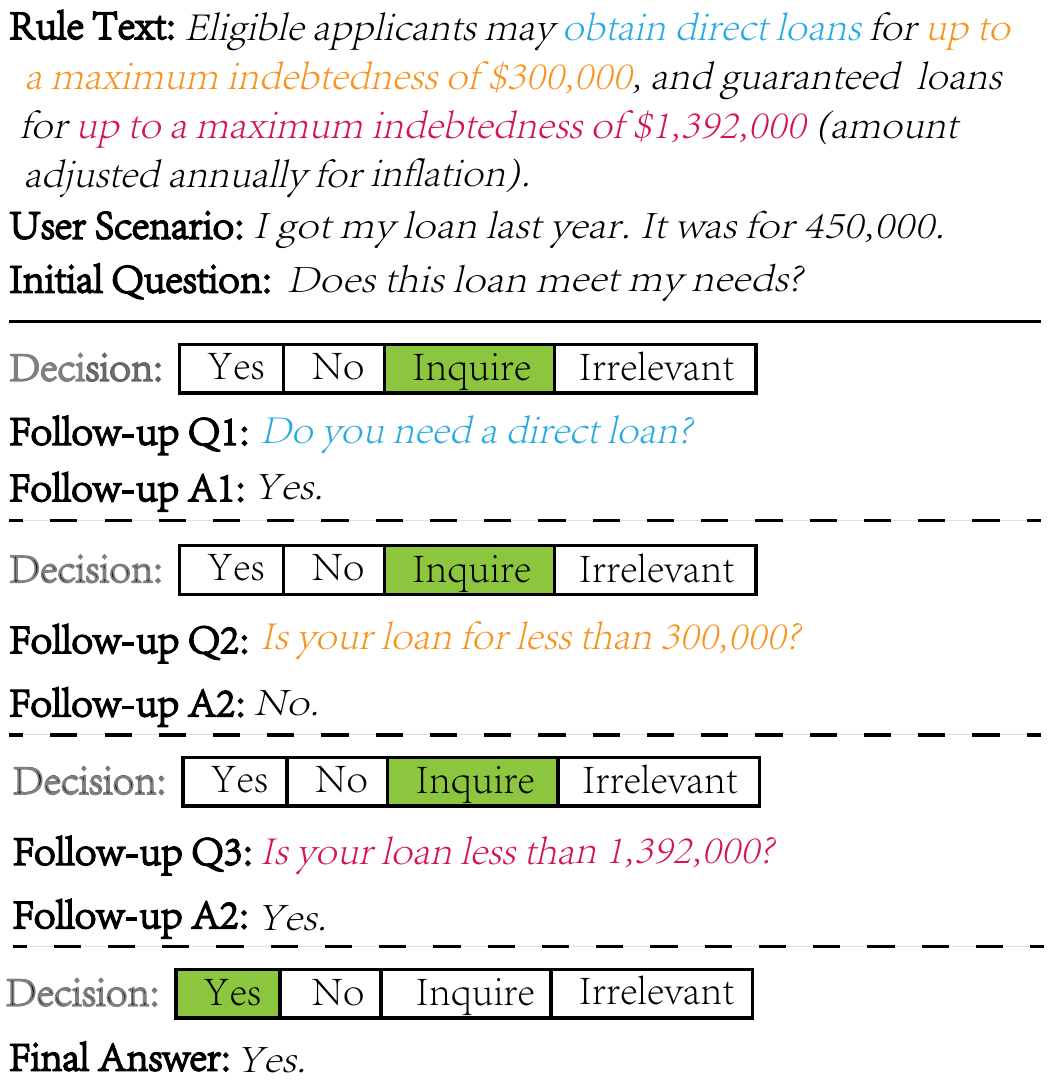}
\caption{An example dialog from ShARC benchmark dataset \citep{saeidi-etal-2018-interpretation}. At each turn, the machine can give a decision regarding the initial question put up by the user. If the decision is \textsl{Inquire}, the machine will ask a clarification question to help with decision making. The corresponding rule document and the question are marked in the same color in the figure.}
\label{sharc-example}
\end{figure}

Training machines to understand documents is the major goal of machine reading comprehension (MRC) \cite{hermann2015teaching,hill2015goldilocks,Rajpurkar2016SQuAD,Nguyen2016MS,Joshi2017TriviaQA,Rajpurkar2018Know,choi2018quac,zhang2018dua,reddy2019coqa,zhang2020mrc,zhang2021retro}. Especially, in the recent challenging conversational machine reading (CMR) task, the machine is required to read and interpret the given rule document and the user scenario, ask clarification questions, and then make a final decision \citep{saeidi-etal-2018-interpretation}. As an example shown in Figure \ref{sharc-example}. The user posts the scenario and asks a question concerning whether the loan meets the needs. Since the user cannot know the rule document, the information he/she provided may not be sufficient for the machine to decide. Therefore, a series of follow-up questions are asked by the machine until it can finally make a conclusion.



The major challenges for the conversational machine reading include the rule document interpretation, and reasoning with the background knowledge, e.g., the provided rule document, user scenario and the input question. Existing works \cite{zhong-zettlemoyer-2019-e3,lawrence-etal-2019-attending,verma-etal-2020-neural,gao-etal-2020-explicit,gao-etal-2020-discern} have made progress in improving the reasoning ability by modeling the interactions among rule document, user scenario and the other elements implicitly. As for rule document interpretation, most existing approaches simply split the rule document into several rule conditions to be satisfied. In general, they first track the entailment state of each rule condition for decision making and then form a certain under-specified rule span into a follow-up question.

However, the aforementioned cascaded methods tend to model in a holistic way, i.e. interpreting the rule document with other elements quite plainly, which have the following drawbacks. First, very little attention is paid to the inner dependencies of rule conditions such as the discourse structure and discourse relations \cite{qin2016stacking,qin2017adversarial,bai2018deep}. Second, existing methods do not dig deep enough into mining the interactions between the rule document and other elements, especially user scenarios. 

As seen, the interactions of elements in CMR is far more complicated than that of traditional MRC tasks. Therefore, we proposed a dialogue graph modeling (DGM) framework consisting of two complementary graphs to fully capture the complicated interactions among all the elements. Firstly, an explicit discourse graph is constructed by making use of discourse relations of elementary discourse units (EDUs) generated from rule documents to tackle explicit element interactions. User scenario representation is injected as a special global vertex, to bridge the interactions and capture the inherent dependency between the rule document and the user scenario information. Secondly, an implicit discourse graph is designed for digging the latent salient interactions among rule documents by decoupling and fusing mechanism. The two dialogue graphs compose the encoder of our model and feed fusing representations to the decoder for making decisions. 


As to our best knowledge we are the first to explicitly model the relationships among rules and user scenario with Graph Convolutional Networks (GCNs) \citep{10.1007/978-3-319-93417-4_38}. Experimental results show that our proposed model outperforms the baseline models in terms of official evaluation metrics and achieves the new state-of-the-art results on ShARC, the benchmark dataset for CMR \citep{saeidi-etal-2018-interpretation}. In addition, our model enjoys strong interpretability by modeling the process in an intuitive way.

\section{Related Work}
\paragraph{Conversational Machine Reading. }
Compared with traditional triplet-based MRC tasks that aim to answer questions by reading given document \cite{hermann2015teaching,hill2015goldilocks,Rajpurkar2016SQuAD,Nguyen2016MS,Joshi2017TriviaQA,Rajpurkar2018Know,zhang2020semantics,zhang2020sg}, our concerned CMR task \citep{saeidi-etal-2018-interpretation} is more challenging as it involves rule documents, scenarios, asking clarification question, and making a final decision. The major differences lie in two sides: 1) machines are required to formulate follow-up questions for clarification before confident enough to make the decision, 2) machines have to make a question-related conclusion by interpreting a set of complex decision rules, instead of simply extracting the answer from the text. Existing works \cite{zhong-zettlemoyer-2019-e3,lawrence-etal-2019-attending,verma-etal-2020-neural,gao-etal-2020-explicit,gao-etal-2020-discern} have made progress in improving the reasoning ability by modeling the interactions between the rule document and other elements. As a widely-used manner, the existing models commonly extracted the rule documents into individual rule items, and track the rule fulfillment for the dialogue states. As indicated in \citet{gao-etal-2020-discern}, improving the rule document representation remains a key factor to the overall model performance, because the rule documents are formed with a series of implicit, separable, and possibly interrelated rule items that the conversation should satisfy before making decisions. However, previous work only considered segmenting the discourse, and neglected the inner discourse structure/relationships between the EDUs \citep{gao-etal-2020-discern}. Compared to existing methods, our method makes the first attempt to explicitly capture elaborate interactions among all the document elements, user scenarios and dialogue history updates.
\paragraph{Graph Modeling in MRC. }Inspired by the impressive performance of GCN \citep{kipf2017semisupervised,luo-zhao-2020-bipartite}, efforts towards better performance on MRC utilizing GCNs have sprung up, such as BAG \citep{cao-etal-2019-bag}, GraphRel \citep{fu-etal-2019-graphrel} and social information reasoning \citep{li-goldwasser-2019-encoding}. Unlike the previous works who just apply the graph framework mechanically to turn the entire passage or document into a graph, the discourse graph we proposed is delicately designed to mine the relationships of multiple elements in CMR task and to facilitate information flow over the graph.



\section{Model}
\begin{figure*}[!t]\label{pipeline}
\centering
\includegraphics[width=1\textwidth]{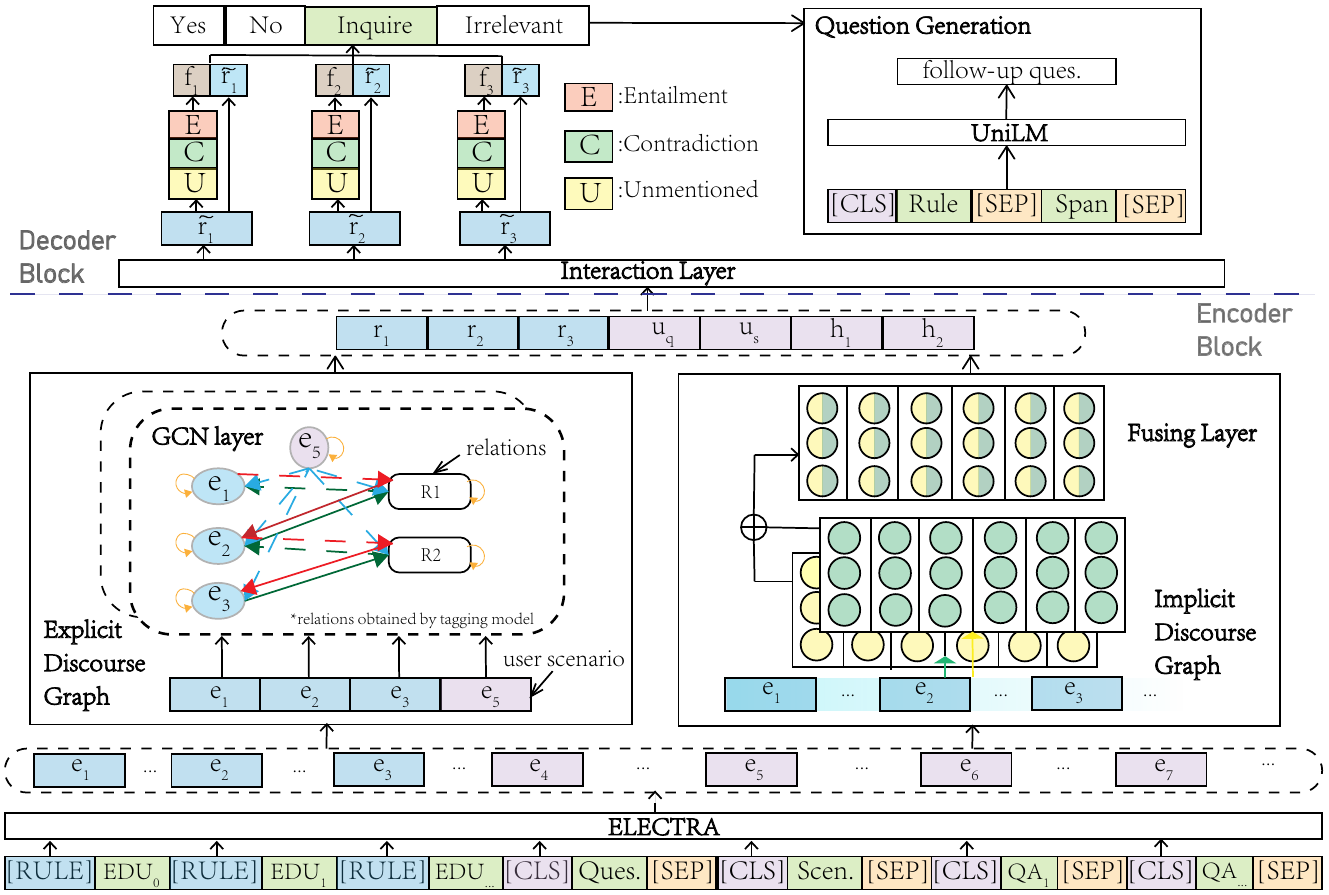}
\caption{The overall structure for our proposed model. With segmented EDUs and tagged relations, the inputs including user initial question, user scenario and dialog history are sent for embedding and graph modeling to make the final decision. If the decision is \textsl{Inquire}, the question generation stage will be activated and use the under-specified span of rule document to generate a follow-up question.}
\label{model-overview}
\end{figure*}

As illustrated in Figure \ref{model-overview}, our model mainly consists of three parts to generate the final answer. 

1.  Rule document is segmented into rule EDUs, which is then tagged discourse relationship by a pre-trained discourse parser.

2.  In the encoding phase, taking segmented and preprocessed rule document and user scenario as input, we build two graphs over the segments (EDUs), in which the explicit discourse graph captures the interactions among rules and user scenarios with the support of tagged discourse relationship, while the implicit discourse graph mines latent salient interactions from the raw rule document.

3.  For decoding, an interaction layer takes the combined representation generated by both explicit and implicit discourse graph of rule EDUs, initial question, user scenario and dialog history as inputs, and maps it into an entailment state of each rule EDU. With these rule fulfillment situation, we can make a decision among \textsl{Yes, No, Inquire and Irrelevant}. Once the decision is made to be \textsl{Inquire}, the model generates a follow-up question to clarify the under-specified rule span in the rule document. 

The complete training procedure of DGM is shown in Algorithm \ref{DGM Algorithm}.

\begin{algorithm}
\caption{DGM Algorithm\label{DGM Algorithm}}
\KwIn{word embeddings $E=\{e_1,..,e_n\}$, dimension of word embeddings $d$, token ids of rule document $I$, discourse relation $D$ of rule document, number of rule EDUs $n$}
\KwOut{Final decision in Yes/No/Irrelevant or a follow-up question}
\BlankLine
\For{$i$ in epochs}
{
\text{build explicit discourse graph $G(E,D)$} \\
$G_{n\times d}\leftarrow GCN(G)$\\
build implicit discourse graph by calculating adjacent matrix $M_l$, $M_c$\\
get rule EDU representation $C_{n\times d}$ by Eq.(\ref{6}) and (\ref{sys})\\
\text{combined representation for \texttt{[RULE]}}$\tilde{r_i}\leftarrow \textup{self-Attn}(C+G)_i$\\
\text{entailment state }$f_i\leftarrow \textup{LINEAR}(\tilde{r_i})$\\
\text{make the decision $z$ by Eq.(\ref{dec})} based on $\tilde{r_i}$ and $f_i$\\
\If{$z$ is \textsl{Inquire}}{\text{generate follow-up question}}
\Return $z$ or follow-up question
}
\end{algorithm}

\subsection{Preprocessing}\label{encoding}
\paragraph{EDU Segmentation.} We first separate the rule document into several units each containing exactly one condition. Here we follow \textsc{Discern} \citep{gao-etal-2020-discern} adopting the discourse segmenter \citep{ijcai2018-579} to break the rule document into EDUs.
\paragraph{Discourse Relation.} Unlike EDU segmentation which only concerns with constituency-based logical structures, discourse relation allows relations between the non-adjacent EDUs. There are in total $16$ discourse relations according to STAC \citep{asher-etal-2016-discourse}, namely, \textit{comment, clarification-question, elaboration, acknowledgement, continuation, explanation, conditional, question-answer, alternation, question-elaboration, result, background, narration, correction, parallel and contrast}. We adopt a pre-trained discourse parser \citep{shi2019deep}\footnote{This discourse parser gives a state-of-the-art performance on STAC so far} to decide the dependencies between EDUs and the corresponding relation types with the structured representation of each EDU.

\subsection{Encoding Block}\label{graph modeling}

\paragraph{Embedding.}
We select the pre-trained language model (PrLM) model ELECTRA \citep{clark2020electra} for encoding. As shown in the figure, the input of our model includes rule document which has already be parsed into EDUs with explicit discourse relation tagging, user initial question, user scenario and the dialog history. Instead of inserting a \texttt{[CLS]} token before each rule EDU to get a sentence-level representation, we use \texttt{[RULE]} which is proved to enhance performance \citep{lee-etal-2020-slm}. Note that we also insert \texttt{[SEP]} between every two adjacent utterances.

\begin{figure*}[h]
\centering
\subfigure[separated EDUs]{\includegraphics[width=0.32\textwidth]{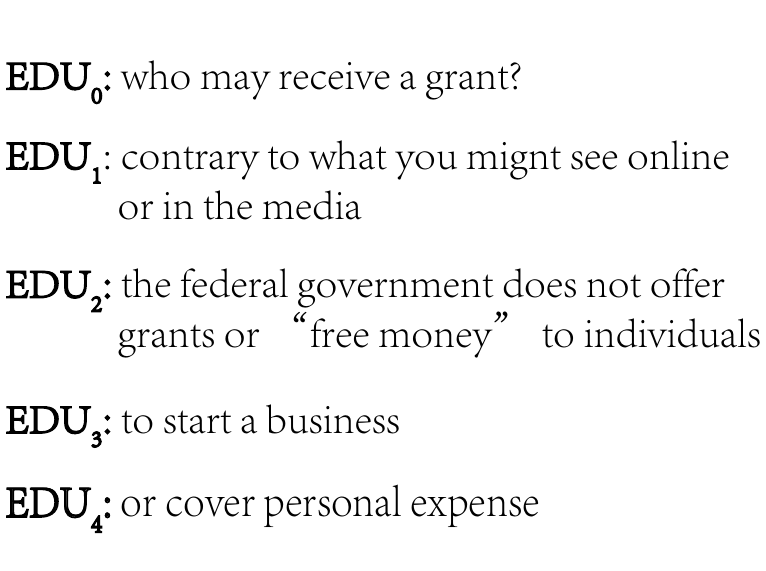}\label{subfigure:EDUs}}
\subfigure[the original graph]{\includegraphics[width=0.32\textwidth]{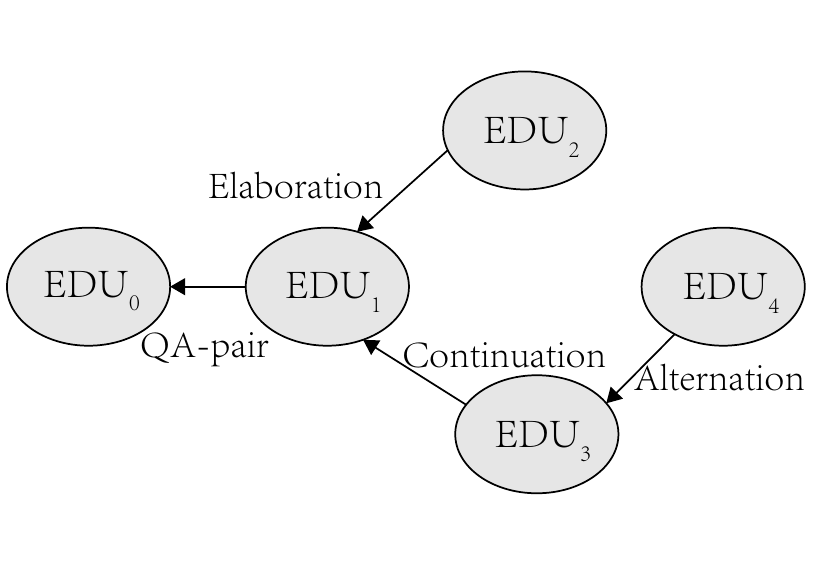}\label{subfigure:original}}
\subfigure[the Levi graph]{\includegraphics[width=0.32\textwidth]{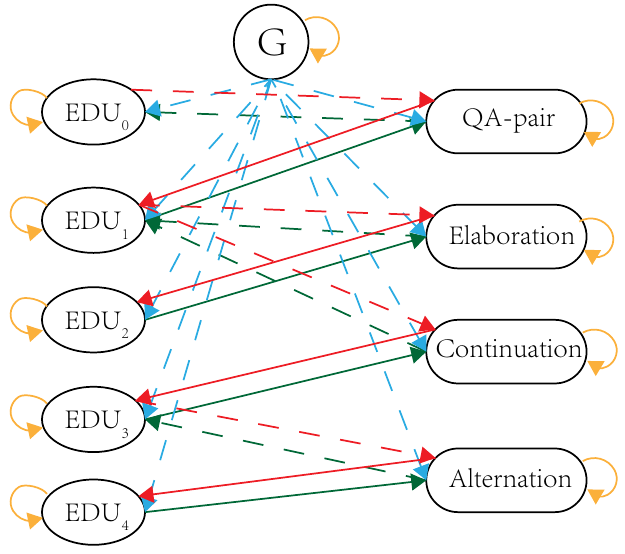}\label{subfigure:levi}}
\caption{Processes turning a sample dialog into Levi graph representing discourse relations.}
\label{fig:discourse graph}
\end{figure*}

\paragraph{Explicit Discourse Graph.}We first construct the explicit discourse graph as a Levi graph \citep{levi-1942} which turns the labeled edges into additional vertices. Suppose $G=(V,E,R)$ is the graph constructed in the following way: if utterance $U_1$ is the continuation of utterance $U_2$, we add a directed edge $e=(U_1,U_2)$ with relation $R$ assigned to \textit{Continuation}. The corresponding Levi graph can be expressed as $G=(V_L,E_L,R_L)$ where $V_L=V\cup R$. $E_L$ is the set of edges with format $(U_1,Continuation)$ and $(Continuation,U_2)$. As for $R_L$, previous works such as \citep{marcheggiani-titov-2017-encoding, beck-etal-2018-graph} designed three types of edges $R_L={default, reverse, self}$ to enhance information flow. Here with our settings, we extend it into six types: \textit{default-in}, \textit{default-out}, \textit{reverse-in}, \textit{reverse-out}, \textit{self}, \textit{global}, corresponding to the direction of the edges towards the relation vertices. An example of constructing Levi graph is shown in Figure \ref{fig:discourse graph}. To construct the discourse structure of other elements, a global vertex representing user scenario is added and connected with all the other vertices.

We use a relational graph convolutional network \citep{10.1007/978-3-319-93417-4_38} to implement explicit discourse graph as the traditional GCN is not able to handle multi-relation graphs. For utterance and scenario vertices, we employ the encoding results of \texttt{[RULE]} and \texttt{[CLS]} in Section \ref{encoding}. For relation vertices, we look up in the embedding table to get the initial representation. Given the initial representation $h_p^0$ of every node $v_p$, the feed-forward or the message-passing process can be written as:
\begin{equation}
    h_p^{(l+1)}=\textup{ReLU}(\sum_{r\in R_L}\sum_{v_q\in \mathcal{N}_r(v_p)}\frac{1}{c_{p,r}}w_r^{(l)}h_q^{(l)}),
\end{equation}
where $\mathcal{N}_r(v_p)$ denotes the neighbors of node $v_p$ under relation $r$ and $c_{p,r}$ is the number of those nodes. $w_r^{(l)}$ is the learnable parameters of layer $l$.

Because the total $16$ relations cannot be treated equally, e.g. relation \textit{Contrast} is much more important than the relation \textit{Continuation}, we introduce the gating mechanism \citep{marcheggiani-titov-2017-encoding}. The basic idea is to calculate a value between $0$ and $1$ for information passing control.
\begin{equation}
    g_{p}^{(l)}=\textup{Sigmoid}(h_p^{(l)}W_{r,g}),
\end{equation}
where $W^{(l)}_{r,g}$ is a learnable parameter under relation type $r$ of the $l$-th layer. Finally, the forward process of gated GCN can be represented as:
\begin{equation}
        h_p^{(l+1)}=\textup{ReLU}(\sum_{r\in R_L}\sum_{v_q\in \mathcal{N}_r(v_p)}g_{q}^{(l)}\frac{1}{c_{p,r}}w_r^{(l)}h_q^{(l)}),
\end{equation}
\paragraph{Implicit Discourse Graph. }Implicit discourse graph aims at digging the salient latent interactions inside rule document. Each token $i$ in rule EDU is represented as a vertex in the graph. We use adjacent matrices to express implicit discourse graph. Two types of matrices $M_l$ and $M_c$ are introduced standing for local and contextualized information:
\begin{equation}
M_l[i,j] = \left\{
            \begin{array}{lcl}
            {0} &\text{if\ } I_i=I_j, & \\
            {-\infty} &\text{otherwise}. &  
            \end{array}  
       \right.
\end {equation}
\begin{equation}
M_c[i,j] = \left\{
            \begin{array}{lcl}
            {0} &\text{if\ } I_i\neq I_j, & \\
            {-\infty} &\text{otherwise}. &  
            \end{array}  
       \right.
\end {equation}
where $I_i$ is the index of token $i$ in EDU. Thus the information containing in rule document are decoupled in two separate aspects. Using multi-head self-attention to encode the graph and denote the length of the whole rule document as $s$, embedding dimension as $d$, we will get the following:
\begin{equation}\label{6}
    G_i=\textup{MHSA}(E,M_i),\ i\in \{l,c\},
\end{equation}
where $G_i\in \mathbb{R}^{s\times d}$ and $E$ is the embedding result from PrLM. \textup{MHSA} denotes the multi-head self-attention \cite{NIPS2017_3f5ee243}.

After enough interactions inside rule EDUs, we then fuse the information \citep{Liu2020FillingTG} of these two implicit discourse graphs-like items\footnote{Taking self-attention weights as edges connecting representations (as node), it can be seen as graph as well.} above in a gated manner by considering both the original and graph encoding representation of rule document.

\begin{equation}
\begin{aligned}
\tilde{E_1}&=\textup{ReLU}(\textup{FC}([E,G_l, E-G_l, E\odot G_l])), \\
\tilde{E_2}&=\textup{ReLU}(\textup{FC}([E,G_c, E-G_c, E\odot G_c])) ,\\
g&=\textup{Sigmoid}(\textup{FC}([\tilde E_1,\tilde E_2])), \\
C&=g\odot G_l+(1-g)\odot G_c,
\end{aligned}
\label{sys}
\end{equation}
where FC is the fully-connected layer and $C\in \mathbb{R}^{s\times d}$. We take the calculated result of the original \texttt{[RULE]} to stand for the updated rule EDUs from $C$, denoted as $c_i$.

\begin{table*}
\small
\centering\centering\setlength{\tabcolsep}{5.0pt}
\begin{tabular}{lcccccccc}
\toprule
\multirow{3}{*}{Model} &
\multicolumn{4}{c}{Dev Set} & \multicolumn{4}{c}{Test Set}\\
&\multicolumn{2}{c}{Decision Making} & \multicolumn{2}{c}{Question Gen.} & \multicolumn{2}{c}{Decision Making} & \multicolumn{2}{c}{Question Gen.}\\
\cmidrule{2-5}
\cmidrule{6-9}
 & Micro & Macro & BLEU1 & BLEU4 & Micro & Macro & BLEU1 & BLEU4 \\ 
\midrule
NMT \citep{saeidi-etal-2018-interpretation}  &-&-&-&-& 44.8 & 42.8 & 34.0 & 7.8\\
CM \citep{saeidi-etal-2018-interpretation}&-&-&-&-  & 61.9 & 68.9 & 54.4 & 34.4\\
BERTQA \citep{zhong-zettlemoyer-2019-e3}   &68.6&73.7&47.4&54.0  & 63.6   & 70.8   & 46.2 & 36.3 \\
UcraNet \citep{verma-etal-2020-neural} &-&-&-&- & 65.1&71.2&60.5&46.1\\
BiSon  \citep{lawrence-etal-2019-attending} &66.0&70.8&46.6&54.1&66.9&71.6&58.8&44.3 \\
E$^3$ \citep{zhong-zettlemoyer-2019-e3}  &68.0&73.4&67.1&53.7 & 67.7  & 73.3    & 54.1 & 38.7  \\
EMT \citep{gao-etal-2020-explicit}&73.2&78.3&67.5&53.2 &69.1 &74.6&63.9&\textbf{49.5}\\
\textsc{Discern} \citep{gao-etal-2020-discern} &74.9&79.8&65.7&52.4&73.2&78.3 & \textbf{64.0}&49.1 \\
\cdashline{1-9}
DGM (ours) & \textbf{78.6}&\textbf{82.2}&\textbf{71.8}&\textbf{60.2}&\textbf{77.4}&\textbf{81.2} & 63.3&48.4\\
\bottomrule
\end{tabular}
\caption{Results on the blind held-out test set and the dev set of ShARC end-to-end task. Micro and Macro stand for Micro Accuracy and Macro Accuracy respectively.}\label{table:e2e}
\end{table*}

\subsection{Decoding Block}
\paragraph{Interaction Layer. }We use an interaction layer to attend to all available information so far to learn in a systematic way. A self-attention layer \citep{NIPS2017_3f5ee243} is adopted here allowing all the rule EDUs and other elements to attend to each other. Let $[r_1,r_2,...;u_q;u_s;h_1,h_2,...]$ denote all the representations, $r_i$ is the combined sentence-level representation of explicit and implicit discourse graph, $u_q$, $u_s$ and $h_i$ stand for the representation of user question, user scenario and dialog history respectively. After encoding, the output can be displayed as $[\tilde{r_1},\tilde{r_2},...;\tilde{u_q},\tilde{u_s};\tilde{h_1},\tilde{h2}...]$.

\paragraph{Decision Making. }Similar to existing works \citep{zhong-zettlemoyer-2019-e3, gao-etal-2020-explicit, gao-etal-2020-discern}, we apply an entailment-driven approach for decision making. A linear transformation tracks the fulfillment state of each rule EDU among Entailment, Contradiction and Unmentioned. At last, the decision is made according to:
\begin{equation}
    f_i=W_f\tilde{r_i}+b_f\in \mathbb{R}^{3},
\end{equation}
where $f_i$ is the score predicted for the three labels of the $i$-th condition. This prediction is trained via a cross entropy loss for multi-classification problems:
\begin{equation}
    \mathcal{L}_{entail}=-\frac{1}{N}\sum_{i=1}^{N}\log \textup{softmax}(f_i)_r,
\end{equation}
where $r$ is the ground-truth state of fulfillment.

After obtaining the state of every rule, we are able to give a final decision towards whether it is \textsl{Yes, No, Inquire} or \textsl{Irrelevant} by attention.
\begin{equation}
\begin{aligned}
    \alpha_i&=w_{\alpha}^T[f_i;\tilde{r_i}]+b_\alpha \in \mathbb{R}^1,\\
    \tilde{\alpha_i}&=\textup{softmax}(\alpha)_i\in [0,1],\\
    z&=W_z\sum_i\tilde{\alpha_i}[f_i;\tilde{r_i}]+b_z\in \mathbb{R}^4,
\end{aligned}
\label{dec}
\end{equation}
where $\alpha_i$ is the attention weight for the $i$-th decision and $z$ has the score for all the four possible states. The corresponding training loss is:
\begin{equation}
    \mathcal{L}_{decision}=-\log \textup{softmax}(z)_l,
\end{equation}
The overall loss for decision making is:
\begin{equation}
    \mathcal{L} =\mathcal{L}_{decision} + \lambda\mathcal{L}_{entail}.
\end{equation}

\paragraph{Qustion Generation. }If the decision is made to be \textsl{Inquire}, the machine need to ask a follow-up question to further clarify. Question generation in this part is mainly based on the uncovered information in the rule document, and then that information will be rephrased into a question. We predict the position of an under-specified span within a rule document in a supervised way. Following \citet{devlin-etal-2019-bert}, our model learns a start vector $w_s\in \mathbb{R}^d$ and end vector $w_e\in \mathbb{R}^d$ to indicate the start and end positions of the desired span:
\begin{equation}
    span=\mathop{\arg\min}_{i,j,k} (w_s^Tt_{k,i}+w_e^Tt_{k,j}),
\end{equation}
where $t_{k,i}$ denote the $i$-th token in the $k$-th rule sentence. The ground-truth span labels are generated by calculating the edit-distance between the rule span and the follow-up questions. Intuitively, the shortest rule span with the minimum edit-distance is selected to be the under-specified span. Finally, we concatenate the rule document and the predicted span as an input sequence to finetune UniLM \citep{NEURIPS2019_c20bb2d9} and generate the follow-up question.

\section{Experiments}
\subsection{Experimental Setup} \label{sec:experiment}
\paragraph{Dataset. }We conduct experiments on ShARC dataset, the current CMR benchmark\footnote{Leaderboard can be found at website \url{ https://sharc-data.github.io/leaderboard.html}} collected by \citet{saeidi-etal-2018-interpretation}. It contains up to 948 dialog trees clawed from government websites. Those dialog trees are then flattened into 32,436 examples consisting of \textit{utterance\_id}, \textit{tree\_id}, \textit{rule document}, \textit{initial question}, \textit{user scenario}, \textit{dialog history}, \textit{evidence} and the \textit{decision}. It is worth noting that evidence is the information that we need to extract from user information and thus will not be given in the testing phase. The sizes of train, dev and test are 21,890, 2,270 and 8,276 respectively. We also showed the generalizability of our model on the Multi-Turn Dialogue Reasoning (MuTual) dataset  \citep{MuTual}, which has 8,678 multiple choice samples and is divided into 7,376, 651, 651 of train, dev and test sets respectively. 
\paragraph{Evaluation. }For the decision-making subtask, ShARC evaluates the Micro- and Macro- Acc. for the results of classification. If both the prediction and ground truth of decision is \textsl{Inquire}, BLEU\citep{papineni-etal-2002-bleu} score (particularly BLEU1 and BLEU4) will be evaluated on the follow-up question generation subtask.
\paragraph{Implementation Details. }For rule EDU relation prediction, we keep all the default parameters of the original discourse relation parser\footnote{\url{https://github.com/shizhouxing/DialogueDiscourseParsing}}, with $F1$ score achieving $55$. In the decision-making stage, we finetune an ELECTRA-based model. The dimension of hidden states is $1024$ for both the encoder and decoder. The training process uses Adam \citep{2014arXiv1412.6980K} for $5$ epochs with learning rate set to $5$e-$5$. We also use gradient clipping with a maximum gradient norm of $2$, and a total batch size of $16$. In the question generation stage, for the sake of consistency, we also use an ELECTRA-based model for span extraction. For UniLM, we finetune it with a batch size of $16$, a learning rate of $2$e-$5$ and beam size is set to $10$ for inference. It takes 3-4 hours for training on a single TITAN RTX 2080Ti GPU (24GB memory).

\subsection{Results}

Table \ref{table:e2e} shows the results of DGM and all the baseline models for the End-to-End task on the blind held-out test set of ShARC\footnote{As indicated in \citep{gao-etal-2020-explicit, gao-etal-2020-discern}, the question generation results normally suffer from randomness. As the focus of this task is the decision making task like previous studies.}. Evaluating results indicate that DGM outperforms the baselines in most of the metrics. In particular, DGM outperforms the previous state-of-the-art model \textsc{Discern} by $4.2$\% in Micro Acc. and $2.9$\% in Macro Acc.

To test the generality of DGM on other different PrLMs and to do a fair comparison with previous models, We alter the underlying PrLMs to other variants in DGM and the previous state-of-the-art model \textsc{Discern} respectively. The results on the dev set of ShARC are shown in Table \ref{different_prlm}. In the first place, DGM performs better than \textsc{Discern} on all the PrLMs, which indicates the all-round superiority of DGM. Additionally, results on ELECTRA is generally better than that of BERT and RoBERTa. This indicates that ELECTRA is an even better trained PrLM. By the aforementioned analysis, our DGM can generally perform well on widely-used PrLMs.

\begin{table}[H]
\small
\centering\centering\setlength{\tabcolsep}{4.5pt}
\begin{tabular}{lcccc}
\toprule
\multirow{2}{*}{PrLMs}  & \multicolumn{2}{c}{ Micro Acc.} & \multicolumn{2}{c}{Macro Acc.}\\
 & \textsc{Discern} & DGM & \textsc{Discern} & DGM \\ 
\midrule
BERT$_{base}$ & 69.8 & 70.4 & 75.3 &76.0  \\
RoBERTa$_{base}$  & 74.9 & 75.8&79.8  & 80.2 \\
ELECTRA$_{base}$  & 75.2 & 75.5 & 79.7 &80.4  \\
BERT$_{large}$ &  72.8&73.0 &77.8 &78.0 \\
RoBERTa$_{large}$ & 76.1  & 76.6 & 80.6 & 81.0 \\
ELECTRA$_{large}$ & 77.2&78.6 &80.3 &82.2\\
\bottomrule
\end{tabular}
\caption{Performance of \textsc{Discern} and DGM on different PrLMs on the dev set of ShARC.}
\label{different_prlm}
\end{table}

In addition, Table \ref{table:class-wise} lists the class-wise classification accuracy of our model. Results demonstrate that our model performs quite satisfactorily for all classification subtasks, outperforming all other models in three of all four subtasks though a minor behind on the \textsl{Irrelevant} subtask. Compared to competent models, our model boosts the performance with a great gain to judge whether the user's requirements need further inquiry or are already fulfilled. It is worth noting that the \textsl{Inquire} subtask is the most fundamental one among all subtasks required by the concerned CMR. The superiority of our model for this core subtask shows that our DGM model indeed effectively captures the complicated interactions among all the concerned document rules and scenarios.

\begin{table}[H]\centering\setlength{\tabcolsep}{5.6pt}
\small
\begin{tabular}{lccccc}
\toprule
Models &Total& Yes & No & Inquire & Irrelevant \\ 
\midrule
BERTQA  & 63.6& 61.2   & 61.0   & 62.6 & 96.4   \\
E$^3$  & 68.0&65.9   & 70.6    & 60.5 & 96.4   \\
UrcaNet  &65.9&63.3  &  68.4   &    58.9&95.7  \\
EMT &73.2&70.5 &73.2&70.8&98.6 \\
\textsc{Discern} &75.2&71.9&75.8&73.3&\textbf{99.3} \\
\cdashline{1-6}
DGM (ours) & \textbf{77.8}&\textbf{75.2}& \textbf{77.9} & \textbf{76.3}&97.8\\
\bottomrule
\end{tabular}
\caption{Class-wise decision prediction accuracy on the dev set of ShARC.}
\label{table:class-wise}
\end{table}

\section{Analysis}
\begin{figure*}[!t]
\centering
\includegraphics[width=1\textwidth]{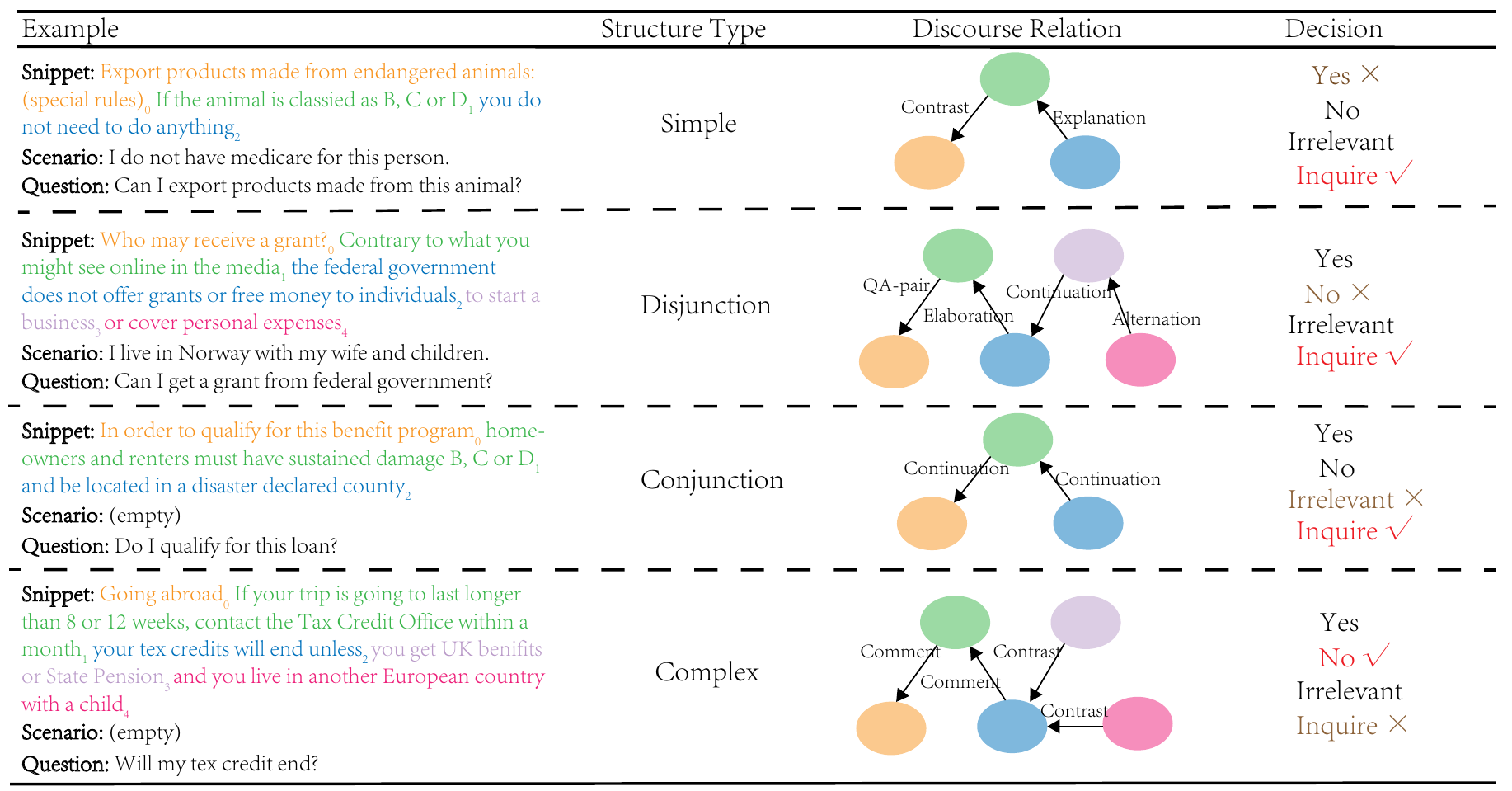}
\caption{Examples selected from the dev set of ShARC where \textsc{Discern} fails but our model succeeds.}
\label{rule_example}
\end{figure*}

\subsection{Ablation Study}

To investigate the impacts of different graphs, we conducted an ablation study on the decision-making subtask which is the vital part of our model, directly influencing the results afterward. Detailed results on the dev set of ShARC in Table \ref{table:ablation} show that both the explicit and implicit discourse graph are indispensable as removing any one of them causes a performance drop ($1$-$3$ points) on both Macro Acc. and Micro Acc. Especially, these two metrics drop by a great margin as we remove the explicit discourse graph, which shows that explicit discourse relation reasoning is crucial in CMR. Also, we can see that adding the special token \texttt{[RULE]} indeed conduce to the performance.

\begin{table}[H]\centering\setlength{\tabcolsep}{5.0pt}
\small
\begin{tabular}{lcc}
\toprule
Models & Macro Acc. & Micro Acc. \\
\midrule
DGM & 82.2 & 78.6 \\
\ w/o Explicit Discourse Graph  & 79.8  & 75.2    \\
\ w/o Implicit Discourse Graph  & 81.3   & 76.7   \\
\ w/o both & 77.3 & 71.7 \\
\cdashline{1-3}
\ w/o [RULE] & 81.6 & 77.9\\
\bottomrule
\end{tabular}
\caption{Ablation study of our model for decision making subtask on the dev set of ShARC.}
\label{table:ablation}
\end{table}

\subsection{User Scenario Interpretation}
In DGM, by injecting the user scenario as the global node in the explicit discourse graph we intend to improve the interpretation ability of our model with respect to user scenarios. To test the effect of the proposed model on scenario interpretation, we create a subset based on the dev set consisting of $761$ samples that have user scenarios and an empty dialog history. The results on the decision-making subtask in Table \ref{table:user_scenario} shows that our model can greatly improve the interpretation of user scenarios by surpassing \textsc{Discern} 11.8\% and 14.8\% of Macro Acc. and Micro Acc. respectively. In particular, DGM outperforms \textsc{Discern} by a large margin in every class of decision.

\begin{table}[H]\centering\setlength{\tabcolsep}{2.6pt}
\small
\begin{tabular}{lcccccc}
\toprule
Models & Macro& Micro &Yes&No&Irrelevant&Inquire \\
\midrule
\textsc{Discern} & 63.5& 60.2 &35.3&50.7&100.0&67.8 \\
DGM & \textbf{75.3} & \textbf{75.0}&\textbf{58.3}&\textbf{62.8}&\textbf{100.0}&\textbf{79.1} \\
\bottomrule
\end{tabular}
\caption{Results for decision making over user scenario subset of the ShARC dev set.}
\label{table:user_scenario}
\end{table}
Manually analyzing the predicted results also indicates that DGM is capable of various reasoning including numerical reasoning, commonsense reasoning and rule document paraphrasing. For example, for numerical reasoning, given a scenario \textit{``I plan on being away for five months before returning"}, DGM is able to match that with \textit{``your tax credits will stop if you expect to be away for one year or more"} in the rule document.

\subsection{Rule Document Interpretation}

To see how DGM interpret the rule document, we analyzed the predictions from DGM and \textsc{Discern} on the dev set of ShARC to see how our model fixes the erroneous cases made by \textsc{Discern}.

We selected four types of rule structures and the representative examples are shown in Figure \ref{rule_example}. It can be seen that the discourse relation tagged for the rule document can well represent the real relation in the discourse.
For example, in the third case, \textit{``homeowners and renters must have sustained damage B, C or D"} is the continuation of \textit{``in order to qualify for this benefit program"} and \textit{"be located in a disaster declared county"} is the continuation of it. This kind of discourse relation informs that \textit{``be located in a disaster declared county"} and \textit{``have sustained damage B, C or D"} are two conditions one must obey to be qualified.
Generally, \textit{Continuation} may indicate that two rules have some conjunctive relations while \textit{Alternation}  denotes a disjunctive relation. All the relations together characterize the complex rule relations and thus are vital in decision making. Statistics regarding relations can be found in Table \ref{relation_total}.
The contextualized information containing in the rule document learned by the implicit discourse graph also contributes to the overall performance as it digs the semantically rich representations of rule EDUs.

\begin{table}[H]\centering\setlength{\tabcolsep}{2.6pt}
\small
\begin{tabular}{lcc}
\toprule
Relation Types & Train Set & Dev Set \\
\midrule
Comment & 28756 & 2374 \\
Clarification\_question & 330 & 69 \\
Elaboration &639&82 \\
Acknowledgement &6242&815 \\
Continuation &7317&1090 \\
Explanation  & 10831&1155\\
Conditional &1445&139 \\
Question-answer\_pair  &1824&468 \\
Alternation &896&323  \\
Result & 664&0 \\
Correction &14 &0 \\
Contrast &16523 &1595 \\
\bottomrule
\end{tabular}
\caption{Statistics analysis of relation types of the train and dev on ShARC. ``Comment", ``Continuation", ``Explanation" and ``Contrast" constitutes the majority of the discourse relations.}
\label{relation_total}
\end{table}

\subsection{Generalizability Evaluation}
To verify DGM's generalizability and show that it can be smoothly applied to a broad type of QA tasks, We conducted experiments on a representative dialogue reasoning dataset MuTual. It is modified from Chinese high school English listening comprehension test data. It consists of 8860 annotated dialogues, namely, 7088 training samples, 886 developing samples and 886 testing samples. For each example, there is a dialogue history following by four candidate responses. Each candidate is relevant to the dialogue context but only one of them is logically correct. Our aim is to predict the correct answer given dialogue history and response candidates.

To apply DGM on MuTual, we first annotated the utterances of dialogue history of their discourse relations. Then pass the dialogue and the response candidates into a pre-trained language model to get the representation of each utterance. Armed with these representations and discourse relations, we are now able to construct the explicit discourse graph. Here, we set the global representation [CLS] \citep{devlin-etal-2019-bert} of dialogue history as the global node. The implicit discourse graph can be constructed as Section \ref{graph modeling} stated.

For the sake of computational efficiency, the maximum number of utterances is set to be 25. The concatenated context, response candidate in one sample is truncated or padded to be of length 256. We use ELECTRA as the PrLM and AdamW \citep{2017arXiv171105101L} as the optimizer for training. The batch size is 24 and the learning rate is 6e-6. We run a total of $3$ epochs and select the model of the best results in the development set.

Table \ref{MuTual_results} displays the results on MuTual, which shows that DGM achieves a consistent improvement on the performance with respect to all the corresponding metrices. 
\begin{table}[H]\centering\setlength{\tabcolsep}{3.5pt}
\small
\begin{tabular}{lcccccc}
\toprule
\multirow{2}{*}{Models}  & \multicolumn{3}{c}{ Dev Set} & \multicolumn{3}{c}{Test Set}\\
 & $R_4@1$ & $R_4@2$ & MRR & $R_4@1$ & $R_4@2$ & MRR \\ 
\midrule
ELECTRA & 90.6 & 97.7 & 94.9 & 90.0 & 97.9 & 94.6 \\
DGM &91.3&98.3 &95.3&90.7&98.2&95.1 \\

\bottomrule
\end{tabular}
\caption{Results on the dev and test set of MuTual dataset}
\label{MuTual_results}
\end{table}

\section{Conclusions}
In this paper, we presented a novel Dialogue Graph Modeling framework for Conversational Machine Reading. Our DGM consists of two complementary graphs which respectively capture both explicit and implicit interactions among multiple complicated elements in the challenging task, in which Explicit Discourse Graph is for extra knowledge learning with tagged EDU discourse relations while Implicit Discourse Graph helps with inside rule document understanding. Experiments on ShARC show the effectiveness by achieving a new state-of-the-art result. Our method may be smoothly applied to a broad type of QA tasks, such as our practice on the MuTual dataset  that also achieves a consistent performance.

\section*{Acknowledgments}
We would like to thank all the anonymous reviewers for their helpful comments and suggestions. Also thanks to Max Bartolo for evaluating our submitted models on the hidden test set.

\bibliographystyle{acl_natbib}
\bibliography{acl2021}

\end{document}